\documentclass[journal]{IEEEtran}

\usepackage{verbatim}
\usepackage{amsfonts}
\usepackage{amssymb}
\usepackage{stfloats}
\usepackage{cite}
\usepackage{graphicx}
\usepackage{psfrag}
\usepackage{amsmath}
\usepackage{array}
\usepackage{epstopdf}
\usepackage{authblk}
\usepackage{graphicx} 
\usepackage{amsthm} 
\usepackage{lipsum}
\usepackage{verbatim} 
\usepackage{authblk}
\usepackage{mathtools}
\usepackage{cuted}
\usepackage{booktabs}

\usepackage{amsmath}
\usepackage{mathrsfs}

\usepackage{algorithmic}

\usepackage{array}

\ifCLASSOPTIONcompsoc
\usepackage[caption=false,font=normalsize,labelfont=sf,textfont=sf]{subfig}
\else
\usepackage[caption=false,font=footnotesize]{subfig}
\fi

\usepackage{url}
\usepackage{graphicx,amsmath,amssymb,amsfonts}
\usepackage{algorithmic,algorithm}

\usepackage{hyperref}
\hypersetup{colorlinks=true, citecolor=blue}

\usepackage{setspace}

\makeatletter

\newcommand{\Rmnum}[1]{\expandafter\@slowromancap\romannumeral #1@}
\makeatother

\usepackage{framed} 

\usepackage{cancel}

\usepackage{booktabs}
\usepackage{tabularx,makecell,multirow}

\usepackage[table]{xcolor}
\usepackage{pifont} 

\usepackage{diagbox}

\begin{document}
	\bstctlcite{ref:BSTcontrol}
	
	\title{Reconfigurable Intelligent Surface for Internet of Robotic Things}
	
	\author{Wanli~Ni, Ruyu Luo, Xinran Zhang, Peng Wang, Wen Wang, and Hui Tian
		
	\thanks{Wanli Ni (corresponding author) is with the Department of Electronic Engineering, Tsinghua University, China (e-mail: niwanli@tsinghua.edu.cn).}
	\thanks{Ruyu Luo, Xinran Zhang and Hui Tian are with the State Key Laboratory of Networking and Switching Technology, Beijing University of Posts and Telecommunications, China (e-mail: luory@bupt.edu.cn, xinranzhang@bupt.edu.cn, tianhui@bupt.edu.cn).}
	\thanks{Peng Wang is with the School of Electrical and Electronic Engineering, North China Electric Power University, Beijing 102206, China (e-mail: wangpeng9712@ncepu.edu.cn).}
	\thanks{Wen Wang is with the Pervasive Communications Center, Purple Mountain Laboratories, Nanjing 211111, China (email: wangwen@pmlabs.com.cn).}
}
	
	\maketitle
	
	\begin{abstract}
		With the rapid development of artificial intelligence, robotics, and Internet of Things, multi-robot systems are progressively acquiring human-like environmental perception and understanding capabilities, empowering them to complete complex tasks through autonomous decision-making and interaction.
		However, the Internet of Robotic Things (IoRT) faces significant challenges in terms of spectrum resources, sensing accuracy, communication latency, and energy supply.
		To address these issues, a reconfigurable intelligent surface (RIS)-aided IoRT network is proposed to enhance the overall performance of robotic communication, sensing, computation, and energy harvesting.
		In the case studies, by jointly optimizing parameters such as transceiver beamforming, robot trajectories, and RIS coefficients, solutions based on multi-agent deep reinforcement learning and multi-objective optimization are proposed to solve problems such as beamforming design, path planning, target sensing, and data aggregation.
		Numerical results are provided to demonstrate the effectiveness of proposed solutions in improve communication quality, sensing accuracy, computation error, and energy efficiency of RIS-aided IoRT networks.
	\end{abstract}

	\section{Introduction}
	The rapid development of artificial intelligence and automation technologies has led to significant advancements in robot systems \cite{Yara2019Survey}.
	These systems are increasingly acquiring human-like capabilities for environmental perception and understanding, enabling them to accomplish complex tasks through autonomous learning, decision-making, and interaction \cite{Yara2019Survey}. 
	The Internet of Robotic Things (IoRT) represents the integration of robotics and the Internet of Things, enabling seamless connectivity, communication, and coordination among a vast array of robotic devices \cite{Luo2024TWC}. 
	The advent of IoRT poses several challenges. For example, spectrum resources are becoming increasingly scarce, sensing accuracy needs to be improved for reliable decision-making, communication latency must be minimized to ensure real-time interactions, and energy supply remains a critical factor given the limited battery life of many low-end robotic devices \cite{Liu2024Reconfigurable, Shu2024RIS, Niotaki2023RF}. Addressing these challenges is crucial for the widespread implementation and effectiveness of IoRT networks.
	In particular, as the number of robotic devices continues to grow, the demand for efficient and reliable communication capabilities increases \cite{Ren2024Connected}.
	However, traditional technologies face challenges in meeting these demands, particularly in complex environments with obstacles and interference.
	
	To address the limitations of current communications with uncontrollable channels, reconfigurable intelligent surfaces (RISs) have emerged as a promising solution, offering the ability to dynamically manipulate electromagnetic waves to optimize communication, sensing, computation, and energy harvesting. However, integrating RISs into IoRT networks presents several challenges, including the need for efficient algorithms for configuring RISs, as well as the position selection of RISs.
	In the literature, the authors of \cite{Liu2024Reconfigurable}, \cite{Luo2022Federated} and \cite{Gao2023Intelligent} proposed the utilization of RISs in IoRT networks to address signal blockage issues, thereby enhancing the performance of wireless communication and robotic motion. Additionally, the authors of \cite{Shu2024RIS} investigated an RIS-aided IoRT network where robots performed target sensing and offloaded computational tasks to the base station (BS).
	%
	However, RISs in the above works are only used to assist robots to offload computing tasks to the BS, ignoring the improvement of sensing performance by RIS.
	Moreover, these existing works lack the exploration of how to improve the wireless charging of robots and overlook the energy efficiency of IoRT networks.
	
	Motivated by this background, in this paper, we investigate an RIS-aided IoRT network that utilizes RISs to enhance the performance from communication and sensing to computation and energy harvesting.
	We then present case studies to show the joint performance optimization in RIS-aided IoRT networks.
	Specifically, the following three cases are studied:
	\begin{itemize}
		\item In the first case, to address the challenges of signal degradation due to obstacles and robot mobility, a federated reinforcement learning algorithm is proposed to achieve intelligent network management. Simulation results show that the proposed algorithm significantly enhances system energy efficiency through intelligent optimization of communication resources and robot trajectories.
				
		\item In the second case, to overcome the issues brought by incomplete sensing information in IoRT networks, a multi-BS cooperative sensing scheme is proposed and a coordinated beamforming design algorithm is designed for the joint optimization of communication and sensing. Simulation results demonstrate the superior performance obtained by the multi-objective optimization method.
		
		\item In the third case, to solve signal distortion due to obstacles and long-distance channel fading, a deep reinforcement learning (DRL) algorithm is proposed to optimize energy harvesting and information transmission. Simulation results demonstrate its effectiveness in handling the uncertainty associated with time-varying channels, while also enabling efficient energy transfer and computation over the air.
	\end{itemize}

	\begin{figure*} [t]
		\centering
		\includegraphics[width=6.0 in]{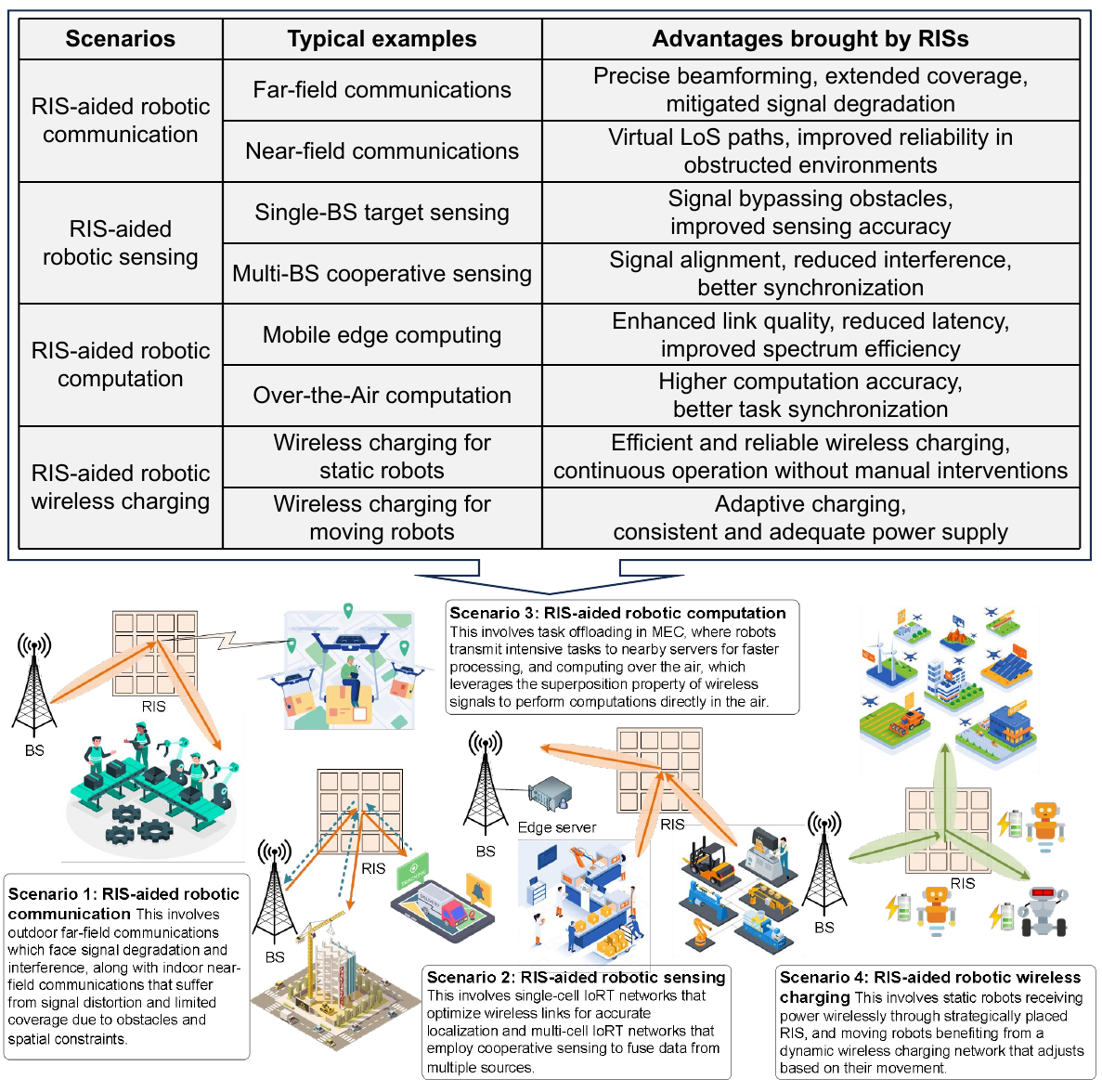}
		\caption{An illustration of RIS-aided robotic communication, sensing, computation, and wireless charging in IoRT networks.}
		\label{Fig1}
	\end{figure*}

	\section{RIS-Aided IoRT Networks}
	
	\subsection{RIS-Aided Robotic Communication}
	Robotic communication faces several challenges, including limited range, signal interference, and high energy consumption.
	To address these challenges, by dynamically adjusting the phase and amplitude of radio waves, RIS can focus signals towards desired locations, thereby extending the communication range and reducing interference \cite{Gao2023Intelligent}.

	\begin{itemize} 
		\item
		\textbf{RIS-Aided Far-Field Robotic Communication:}
		The electromagnetic waves in far-field communication propagate as planar wavefronts, with their phase and amplitude primarily varying based on distance and angle.
		Far-field robotic communication in outdoor scenarios often faces challenges such as signal degradation and interference.
		With the integration of RIS, these challenges can be mitigated, and thus enhancing the performance of outdoor robotic communication scenarios.
		This is particularly beneficial for robots operating in large outdoor environments, such as agricultural fields, construction sites, and surveillance areas. 
		Furthermore, RIS can be used to adapt to dynamic changes in the environment, such as the movement of objects or changes in weather conditions, ensuring continuous and uninterrupted communication.
		
		\item 
		\textbf{RIS-Aided Near-Field Robotic Communication:}
		In contrast, in indoor environments where space is limited, near-field communication may become prevalent.
		In such cases, it is necessary to consider the propagation characteristics of spherical waves as well as their nonlinear phase properties.
		The incorporation of RIS in such scenarios is advantageous as it enables the shaping of electromagnetic waves into elliptical equi-phase surfaces in the near-field range, instead of spherical ones. This offers fresh prospects for achieving precise beamforming and focusing capabilities.
	\end{itemize}

	\subsection{RIS-Aided Robotic Sensing}
	As the complexity and scale of IoRT systems continue to grow, the demand for accurate sensing and localization has become increasingly critical \cite{Wang2024Adaptive}.
	The significance of these capabilities lies in its potential to transform how robots interact with the physical world and each other, leading to more coordinated and intelligent robotic systems.
	However, several challenges persist, including the need for robust solutions that can operate in diverse and dynamic environments.
	By leveraging RIS, we can develop innovative solutions to enhance the sensing and localization capabilities of robots.
	
	\begin{itemize} 
		\item
		\textbf{RIS-Aided Robotic Sensing in Single-Cell Networks:}
		In single-cell IoRT networks, the RIS can be strategically placed to optimize the wireless link between the BS and robots.
		By dynamically adjusting RISs, the signal-to-noise ratio (SNR) of echo signals can be maximized, thereby enhancing the accuracy of sensing and localization.
		Furthermore, the RIS can be used to create multiple virtual paths for signal transmission, providing diversity and robustness against potential disruptions.
		
		\item 
		\textbf{RIS-Aided Cooperative Sensing in Multi-Cell Networks:}
		By utilizing multiple BSs, a more comprehensive and robust sensing strategy can be designed.
		This cooperative approach allows for the distribution of sensing tasks across multiple BSs.
		Furthermore, the cooperative sensing approach allows for data fusion from multiple sources, providing a more comprehensive understanding of the environment. This is particularly important in complex or obstructed environments, where a single BS may not be able to provide sufficient coverage or accuracy, such as underground facilities, urban canyons, or industrial sites with heavy machinery.
	\end{itemize}

	\subsection{RIS-Aided Robotic Computation}
	As robots become more sophisticated and interconnected, they generate vast amounts of data that need to be processed swiftly.
	Traditional cloud-based computing solutions often suffer from high latency.
	Mobile edge computing (MEC) addresses these issues by bringing computational resources closer to the network edge, enabling faster data processing and reduced latency.
	Additionally, over-the-air computation (AirComp) leverages the superposition property of wireless signals to perform data aggregation directly in the air \cite{Ni2022Integrating}, reducing the need for individual data transmission.
	By leveraging RIS, more efficient data processing schemes can be developed to improve the computation performance of robots.
	
	\begin{itemize} 
		\item \textbf{RIS-Aided Robotic Computation Offloading:}
		MEC brings computational resources closer to the source of data, which is particularly beneficial for IoRT networks where latency is a critical concern.
		However, the wireless communication channels between robots and MEC servers can be unreliable, leading to data loss and increased energy consumption.
		RIS-aided MEC not only enhances the reliability and speed of data offloading but also reduces transmission latency, crucial for the seamless execution of complex robotic operations in various environments.
		
		\item \textbf{RIS-Aided Robotic Computation Over The Air:}
		AirComp represents a transformative approach to enable the data aggregation from multiple devices without the need for individual transmission.
		The integration of RIS with AirComp further amplifies its potential by optimizing the wireless channel for efficient data fusion.
		RIS can be used to adjust weighting coefficients of concurrent transmissions, ensuring accurate computation over the air.
	\end{itemize}
	
	\subsection{RIS-Aided Robotic Wireless Charging}
	Traditional robotic charging methods, such as wired connections or replaceable batteries, can be cumbersome and impractical, especially in large-scale deployments or environments where accessibility is limited \cite{Niotaki2023RF}.
	To address these challenges, RIS-aided wireless charging emerges as a promising solution, where robots can be powered in a more efficient and flexible manner.
	The operational range and endurance of robots can be extended, which enables them to perform tasks for longer periods without the need for frequent recharging.
	
	\begin{itemize} 
		\item \textbf{RIS-Aided Wireless Charging for Static Robots:}
		For static robots, such as those used in surveillance or environmental monitoring, RIS-aided wireless charging provides a convenient and reliable power source.
		As such, static robots can operate continuously without the need for frequent battery replacements or manual interventions, significantly improving overall system performance and reducing maintenance costs.
		
		\item \textbf{RIS-Aided Wireless Charging for Moving Robots:}
		For moving robots, such as those used in autonomous navigation or delivery systems, RIS-aided wireless charging presents a more complex challenge due to their dynamic nature and the need for continuous power supply.
		The dynamic beamforming of RISs ensures that moving robots receive consistent and adequate power.
		This not only enhances the operational efficiency of robots but also eliminates the downtime associated with battery replacements or wired charging. 		
	\end{itemize}

	\section{Case Studies in RIS-Aided IoRT Networks}
	
	\begin{figure*} [t]
		\centering
		\includegraphics[width=7 in]{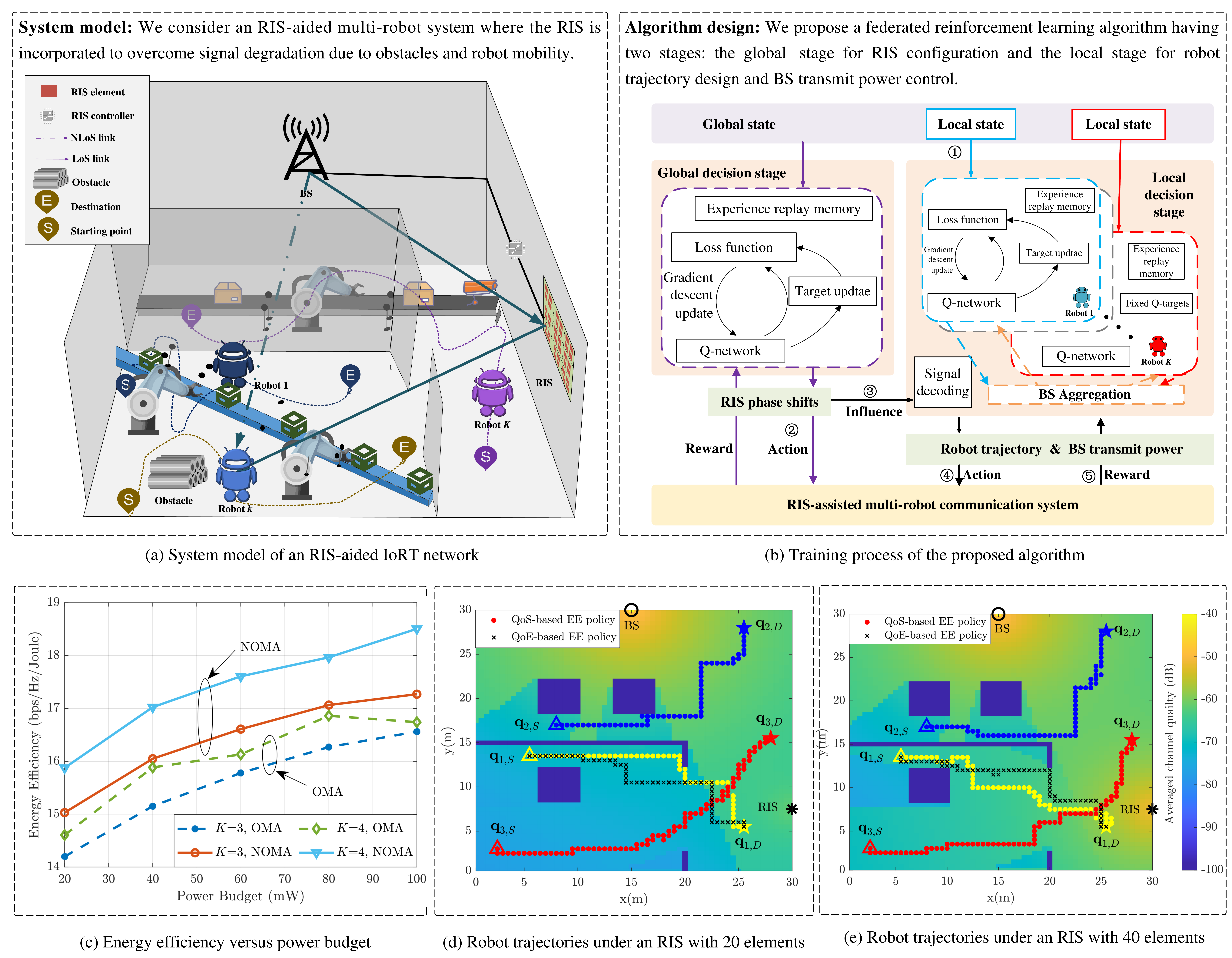}
		\caption{System model, algorithm design, and  simulation results of the joint optimization of communication and trajectory in RIS-aided IoRT networks.}
		\label{Fig2}
	\end{figure*}

	\subsection{Case 1: Joint Design of Communication and Trajectory}
	\subsubsection{Background}
	In an IoRT network, multi-robot communications face severe challenges due to signal degradation caused by obstacles like walls and the dynamic nature of robot mobility.
	Traditional approaches, such as employing active relays, often suffer from scalability and high energy consumption~\cite{Luo2023TCOM}.
	To tackle these issues, RISs are introduced to create smart propagation environments for enhancing signal strength and aiding in trajectory design in a cost-efficient manner.
	While RISs offer substantial advantages in terms of reducing signal blockages and improving communication efficiency, the integration of RIS and IoRT introduces new challenges.
	Specifically, joint optimization of robots' trajectories and RIS configurations is a complex task in a highly dynamic environment.
	Additionally, the fluctuating wireless environment complicates optimal power allocation to maximize sum rates over time-varying channels.
	Traditional methods struggle to optimize resource allocation in RIS-assisted IoRT system, due to the non-convex, integer, and long-term problem formulated.
	Recent research has explored DRL algorithms including centralized or decentralized multi-agent solutions \cite{Ren2024Connected, Gao2023Intelligent}.
	Although decentralized methods offer greater scalability, they can result in conflicts and inefficiencies when robots fail to communicate effectively~\cite{Luo2023TCOM}.
	Therefore, it is necessary to develop new solutions that balance performance, scalability, and adaptability in dynamic environments.
	
	\subsubsection{System Model}
	As shown in Fig. \ref{Fig2}(a), a multi-robot system is considered where the BS serves a group of mobile robots.
	The environment is cluttered with obstacles and walls that can obstruct the line-of-sight (LoS) channels between the robots and the BS.
	To improve communication efficiency and overcome these obstructions, an RIS is utilized to tailor the propagation conditions of the wireless signals.
	Each robot needs to move from its initial position to a designated destination within a given deadline.
	To avoid the sudden acceleration or direction changes, we set each robot update its path at regular intervals and maintains a steady speed \cite{Luo2022Federated}.
	The challenge is to optimally design the trajectories of all robots and allocate the communication resources to maximize the overall energy efficiency of the multi-robot system. The energy efficiency of a robot is determined by the ratio of the total information successfully communicated to the total energy consumed during its movement. 
	%
	Therefore, the problem is formulated as finding the optimal settings for the RIS phase shifts, the transmit power, and the trajectories of all robots. The objective is to maximize the long-term system energy efficiency, considering the constraints imposed by the limited time, power, and frequency resources available.
	Traditional optimization methods are inadequate for such a dynamic and time-coupled problem, thus necessitating the development of intelligent algorithms to efficiently find optimal solutions.

	\subsubsection{Proposed Solution}
	To solve the long-term energy efficiency maximization problem, we propose a federated deep reinforcement learning (FDRL) algorithm to jointly optimize the communication strategies and the trajectories of robots~\cite{Luo2022Federated}.
	As shown in Fig. \ref{Fig2}(b), the proposed FDRL algorithm consists of two stages: a global stage for configuring the RIS, and a local stage for optimizing the trajectories of robots as well as the transmit power of the BS.
	Specifically, the global stage is modeled as a Markov decision process (MDP) that captures the sequential and decision-dependent dynamics of the system.
	The global state space includes the positions of all robots and the channel coefficients between the BS and each robot. 
	The global action space involves selecting discrete phase shifts for each RIS element.
	The global reward function aims to maximize the system sum rate.
	Similarly, the MDP of each robot is defined with the local state space that includes its own position and the direct channel coefficient with the BS.
	The local action space controls the movement direction and the transmit power.
	The local reward function balances energy efficiency and task completion by combining the data rate, time penalties, and goal bonuses.
	During the training process, the interaction between the global and local stages occurs as follows:
	i) Agents observe the environmental states.
	ii) The BS controls the phase shifts of the RIS and the decoding order based on the global model.
	iii) Each robot selects its action for movement and transmit power based on the local model.
	iv) Agents receive rewards and store transitions for future learning.
	v) At each aggregation step, robots periodically upload their local network weights to the BS. The BS then updates the global model and distributes it back to all robots.	
	Following a semi-distributed paradigm, the proposed FDRL algorithm reduces computational complexity compared to centralized methods by decomposing the high-dimensional optimization problem into multiple simplified sub-problems.

	\subsubsection{Simulation Results}
	To evaluate the effectiveness of the proposed FDRL algorithm, simulations are conducted to observe the energy efficiency and robot trajectory under varying system parameters.
	In Fig. \ref{Fig2}(c), the energy efficiency is evaluated against the maximum transmit power for varying numbers of robots, $K$, under both orthogonal multiple access (OMA) and non-orthogonal multiple access (NOMA) schemes.
	From the figure, it is evident that the energy efficiency generally improves with an increase in the power budget for both OMA and NOMA. This is because a higher power budget allows for stronger signal transmissions, which can lead to better SNRs and, consequently, higher energy efficiency. The figure also shows that NOMA outperforms OMA in terms of energy efficiency across different values of $K$. This superiority can be attributed to the fact that NOMA allows multiple users to share the same time-frequency resources, which can lead to more efficient spectrum utilization and, as a result, improved energy efficiency.
	Furthermore, the energy efficiency gap between NOMA and OMA widens as the number of robots $K$ increases. This trend suggests that the advantage of NOMA becomes more pronounced in scenarios with a higher density of users, where the ability to serve multiple users simultaneously without orthogonal resource allocation becomes increasingly beneficial.
	%
	In Figs. \ref{Fig2}(d) and \ref{Fig2}(e), the trajectories of robots are illustrated.
	Two policies are compared: one based on quality-of-service (QoS) and another based on quality-of-experience (QoE). The background shading in the figures represented the quality of the downlink channels, illustrating how the RIS enhances channel conditions, particularly by mitigating severe signal degradation caused by obstacles such as walls.
	The above simulation results confirm that the RIS not only improves the overall quality of communications and energy efficiency but also facilitates better trajectory planning for the robots.

	\begin{figure*}[t]
		\centering
		\centering
		\includegraphics[width=7 in]{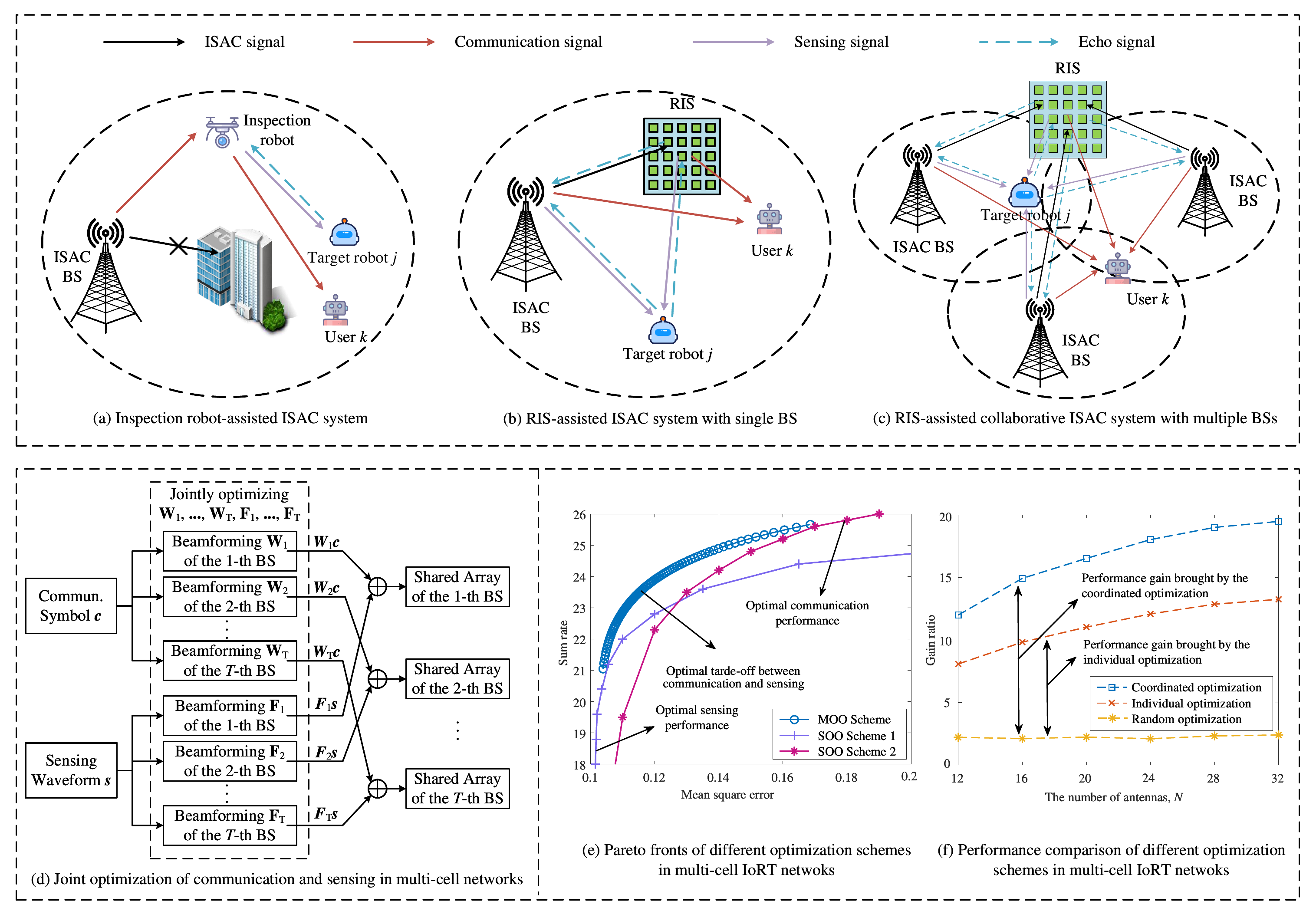}
		\caption{An illustration of the joint optimization of robotic communication and sensing in IoRT networks.}
		\label{Fig3} 
	\end{figure*}	
	
	\subsection{Case 2: Joint Design of Communication and Sensing}	
	
	\subsubsection{Background}
	A lot of intelligent applications in IoRT networks require seamless integration of high-quality communication and high-precision sensing capabilities, driving the development of integrated sensing and communication (ISAC) for robots.
	For example, using ISAC, robots can sense and identify obstacles collaboratively, so as to plan the optimal driving path and avoid collisions.
	Despite its potential, ISAC in IoRT face two main challenges: signal strength degradation due to blockages and incomplete sensing information captured by a single BS.
	The first issue can be mitigated by deploying inspection robots and RIS to aid the transmission of communication and sensing signals, as shown in Fig. \ref{Fig3}(a) and Fig. \ref{Fig3}(b), respectively.
	The second issue can be alleviated by coordinating multiple BSs to capture richer sensing information.
	However, the collaboration of multiple BSs and the introduction of RIS significantly increase the complexity of waveform design.
	Jointly designing active and passive beamforming of BSs and RIS to improve both communication and sensing performance in IoRT networks presents a major
	challenge.
	
	\subsubsection{System Model}
	Based on the coordinated multi-point transmission technology, an RIS-aided ISAC system for IoRT is considered in Fig. \ref{Fig3}(c).
	In this system, multiple ISAC BSs communicate with a number of IoRT users and simultaneously sense multiple IoRT targets.
	The RIS plays a crucial role in overcoming signal blockages by intelligently reflecting signals.
	All ISAC BSs are assumed to be synchronized, transmitting ISAC signals to targets and users.
	Subsequently, they receive echoes from the targets to extract sensing information.	
	Since the beamforming matrices for both communication and sensing signals are designed in a coordinated manner, the main problem in this case focuses on how to achieve the optimal performance in both communication and sensing.
	Traditionally, optimization might focus on either maximizing the sum rate for communication or minimizing the beampattern matching error for sensing \cite{He2023Full}.
	However, the goal here is to strike a balance that achieves optimal performance in both areas, without compromising the requirements of either function.
	The formulated optimization problem includes maximizing the sum rate for communication and minimizing the beampattern matching error for sensing. Constraints are placed on the power levels to ensure feasible solutions. By adjusting the weights representing the decision-maker's preferences, a Pareto front composed of Pareto-optimal solutions can be obtained, indicating the optimal values of communication and sensing under various performance preferences.
	
	\subsubsection{Proposed Solution}
	The proposed solution aims to maximize the communication sum rate and minimize the sensing error by coordinating the beamforming matrices of multiple BSs.
	To this end, the single-objective optimization (SOO) problems are transformed into a multi-objective optimization (MOO) problem using a weighted Tchebycheff-based transformation method.
	Please refer to our prior work in \cite{Wang2024Multi} for more details.
	This method allows for the adjustment of priorities between communication and sensing performance by assigning weight coefficients. By varying these weights, it becomes possible to explore a range of trade-offs between the two objectives, ultimately yielding a Pareto front of solutions that represent the best compromise between communication sum rate and sensing accuracy.
	This approach enables the system to adapt dynamically to different performance preferences, providing flexibility in managing the communication and sensing requirements of IoRT applications. Through this multi-objective approach, the system can be optimized to deliver improved performance in both communication and sensing tasks, addressing the challenges posed by the dynamic and complex nature of IoRT environments.

	\subsubsection{Simulation Results}
	Simulation results in Figs. \ref{Fig3}(e) and~\ref{Fig3}(f) validate the superiority of the proposed MOO algorithm and the coordinated optimization scheme, when the transmit power budget is set as 25 dBm.
	In Fig. \ref{Fig3}(e), the performance boundaries of these SOO schemes are obtained by adjusting the thresholds of the sensing and communication constraints.
	The results indicate that while the SOO schemes deliver better performance in either communication or sensing alone, the proposed MOO scheme provides a better balance between communication and sensing performance.
	In Fig. \ref{Fig3}(f), we consider three BSs, four targets, and six users. The path loss exponent is set as 3.5, the noise power is -80 dBm, the transmit power budget is 20 dBm.
	Our observations reveal that the coordinated optimization scheme exhibits superior system performance compared to the two benchmark schemes.
	This superiority is attributed to the increased spatial degrees of freedom achieved by the joint transmission of signals, facilitated by the coordination among multiple BSs.
	This, in turn, leads to enhanced communication and sensing capabilities.

	\subsection{Case 3: Joint Design of AirComp and Wireless Charging}
	
	\subsubsection{Background}
	The integration of AirComp and wireless charging presents a novel paradigm for enhancing system efficiency and sustainability of robots in IoRT networks.
	Unlike the conventional transmit-then-compute method, AirComp allows multiple robots to transmit data simultaneously, and the BS computes a function of these transmitted signals in the air, thus reducing latency and computational load. Additionally, wireless charging enables battery-limited robots to replenish their energy from radio-frequency (RF) signals, prolonging their operational lifespan \cite{Zhang2024TGCN}.
	However, this integration poses significant challenges. Both downlink wireless charging and uplink AirComp can be adversely affected by channel fading and obstacles in IoRT networks.
	To address these issues, RIS can be utilized to alleviate signal distortion and fading in both downlink and uplink transmissions.
	Despite the benefits, acquiring perfect CSI of all RIS channels poses a significant challenge, especially in dynamic IoRT networks.
	Furthermore, existing beamforming management strategies face difficulties in swiftly adapting to rapidly changing robot environments.
	Consequently, it is crucial to develop an online beamforming design approach that does not rely on prior CSI knowledge.
	In the following, we design AirComp and wireless charging jointly by developing an intelligent algorithm that enables real-time decision-making.

	\begin{figure*} [t]
		\centering
		\includegraphics[width=6.8 in]{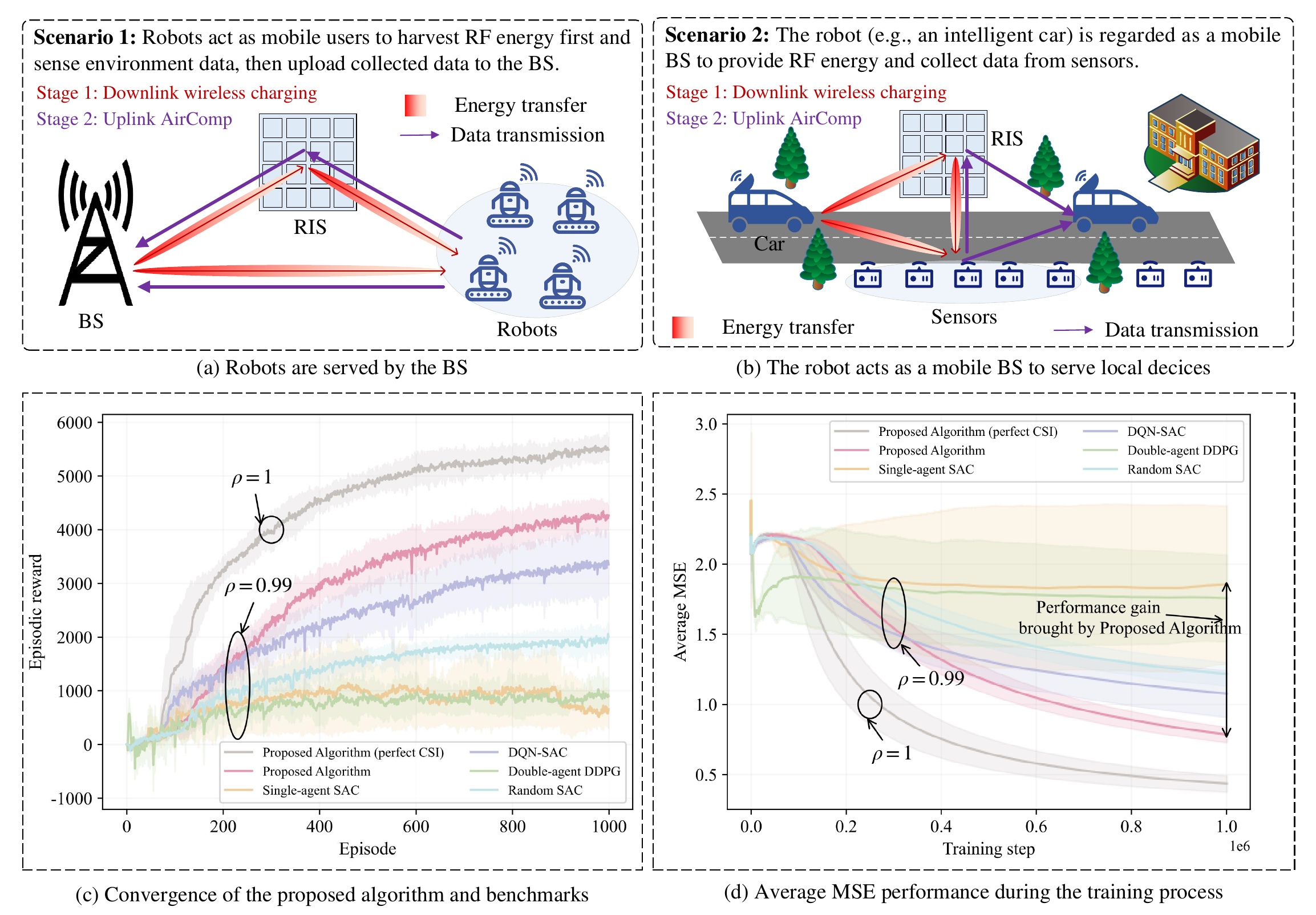}
		\caption{An illustration for the joint optimization of communication, over-the-air computation, and wireless charging in RIS-aided IoRT networks.}	
		\label{Fig4}
	\end{figure*}

	\subsubsection{System Model}
	Fig. \ref{Fig4} shows two scenarios that integrates communication, AirComp, and wireless charging in RIS-aided IoRT networks.
	In Fig. \ref{Fig4}(a), the first scenario considers mobile robots as IoRT users to be first charged by the BS, followed by uplink data transmission.
	In Fig. \ref{Fig4}(b), the second scenario regards robots (e.g., intelligent cars) as mobile BSs to distribute wireless energy and then receive sensing data from local devices. 
	In the following, we take the first scenario as an example to introduce the joint optimization.
	In this case, an RIS is used to assist the downlink wireless charging and uplink AirComp in an IoRT network consisting of a BS and multiple robots.
	The channel uncertainty induced by robot mobility gives rise to the channel aging issue, which is modeled by the temporal correlation coefficient $\rho$. Furthermore, the statistical CSI error due to inaccurate channel estimation is considered at the start of uplink transmissions. 
	The accuracy of AirComp-based data aggregation is measured by the mean square error (MSE), which arises from the factors such as signal misalignment, CSI error, and receive noise.
	The aim of the considered IoRT network is to minimize long-term MSE by jointly optimizing downlink energy beamforming, uplink transceiver beamforming, and downlink/uplink RIS coefficients.
	However, the lack of future dynamic information and the non-convex mixed-integer non-linear programming (MINLP) nature of the problem render it intractable using traditional optimization methods.

	\subsubsection{Proposed Solution}
	To address the challenges posed by time-varying channel uncertainty and the absence of future dynamic information, a double-agent DRL algorithm is proposed \cite{Zhang2024Deep}. This algorithm enables the prediction of future environmental states and the acquisition of system variation rules without prior knowledge. For the proposed DRL algorithm, online decision-making is possible with only the knowledge of imperfectly estimated CSI at the start of the uplink transmission.
	The two agents in the proposed double-agent DRL algorithm share a common state space and a unified reward function, but they operate in distinct action spaces.
	The state space encapsulates the previously recorded MSE value and the imperfectly estimated CSI of all channels at the beginning of the uplink transmission period.
	The reward function is crafted with the dual goals of minimizing the long-term MSE and satisfying specific constraints.
	Agent 1 is responsible for controlling downlink/uplink RIS coefficients.
	Agent 2 performs downlink energy beamforming and uplink transceiver beamforming vectors.
	Furthermore, the proposed double-agent DRL algorithm employs an asynchronous update framework to ensure effective interaction between the two agents.
	Firstly, the future action chosen by Agent 1 serves as a prerequisite for the parameter update of Agent 2. 
	Secondly, the current action executed by Agent 2, in turn, influences the parameter update of Agent 1.	
	In summary, Agent 1's actions act as a condition for Agent 2's network parameter update, and conversely, Agent 2's actions affect Agent 1's network updates. The updates of the two agents alternate asynchronously, with Agent 1 initiating the process followed by Agent 2.
	Together, one update cycle of Agent 1 paired with that of Agent 2 constitutes a single training iteration. 
	This process continues until convergence is achieved or the maximum number of training steps is attained.
	For more detailed information on the convergence and complexity analysis of DRL algorithms, please refer to our previous works in \cite{Luo2024TWC} and \cite{Luo2023TCOM}.

	\subsubsection{Simulation Results}
	In the experiments, we consider a scenario involving a BS and two robots.  The noise power is set to -100 dBm. Additionally, we consider temporal correlation coefficients of $\rho=0.99$ and $\rho=1$, which correspond to robot velocities of $v=1$ m/s and $v=0$ m/s, respectively.
	Regarding the DRL parameters, both agents utilized an identical fully-connected neural network structure with two hidden layers, containing 128 and 32 neurons, respectively. The training process involved 1000 episodes, each consisting of 1000 training steps.
	To validate the performance of the proposed algorithm, several benchmark schemes were considered: 
	1) Random SAC, where Agent 1 generates actions randomly using the soft actor-critic (SAC) algorithm.
	2) Single-agent SAC, employing a single agent with the SAC algorithm to output actions.
	3) Double-agent DDPG, replacing SAC with the deep deterministic policy gradient (DDPG) algorithm, while the double-agent framework is used similar to the proposed algorithm.
	4) DQN-SAC, utilizing a deep Q-network (DQN) by Agent 1 instead of SAC to obtain actions.
	The numerical results in Fig. \ref{Fig4}(c) demonstrate the convergence performance of the proposed algorithm and other DRL-based benchmarks.
	With $\rho$ set to 0.99 for imperfect CSI and 1 for perfect CSI, the proposed algorithm achieves an average MSE closest to the perfect CSI scenario despite some performance loss.
	This indicates that the proposed algorithm is capable of online decision-making even with partial environment knowledge, performing closely to scenarios with perfect information.
	Moreover, in Fig. \ref{Fig4}(d), compared to Single-agent SAC and Random SAC schemes, the proposed algorithm achieves reductions in average MSE of 57.6\% and 35.4\%, respectively.
	Other benchmarks exhibit inferior performance in terms of convergence performance and MSE mitigation.
	For DQN-SAC, the dimensionality curse limits DQN's potential, while Double-agent DDPG's reliance on deterministic policies restricts action exploration, potentially leading to local optimal solutions.
	
	\section{Open Issues and Research Opportunities}
	
	\subsection{Ultra-Reliable Low-Latency Communication for IoRT}
	In IoRT, reliable communication and low latency are crucial for ensuring the safety and operational efficiency of real-time applications. For example, in smart factories, robots operating under tight constraints can experience failures or accidents due to unexpected delays.
	Therefore, ultra-reliable low-latency communication (URLLC) is essential for robotic systems requiring precise control and real-time responsiveness.
	Future research could further explore optimizing IoRT within 6G networks, including intelligent resource allocation, innovative network architecture designs, and communication protocol upgrades. Specifically, it should address communication interference in complex environments by developing robust protocols to ensure the stability and reliability of data transmission.
	Additionally, integrating edge and cloud computing within IoRT can enhance real-time data processing through reduced latency, while leveraging the powerful capabilities of cloud computing to handle complex tasks.
	
	\subsection{Enhancing Security for IoRT}	
	With the growing interconnectivity of IoRT systems, cyber threats have intensified.
	Future research should focus on advanced cybersecurity strategies, including real-time threat response and intrusion detection, to build robust defenses for IoRT networks.
	For example, developing more secure communication protocols is crucial to prevent data leakage during transmissions between robots or between the cloud and robots, effectively countering various cyber attacks. 
	Moreover, strengthening identity authentication mechanisms in distributed and heterogeneous environments will block unauthorized access.
	Given that IoRT systems involve the transmission and processing of large amounts of sensitive data, protecting data privacy is paramount.
	Future research can explore the application of innovative technologies such as differential privacy and federated learning to enable effective data utilization and analysis while safeguarding data privacy.
	
	\subsection{Embodied Intelligence for IoRT}
	A pivotal direction for future IoRT advancements lies in enhancing robots' environmental perception and elevating their autonomous decision-making and action capabilities.
	This necessitates a thorough exploration of the seamless integration of cutting-edge AI technologies, such as computer vision and natural language processing (NLP), with IoRT systems to propel embodied intelligence to new heights.
	Specifically, developing multimodal and multitasking large models that help robots actively collaborate within physical environments is paramount.
	These models will empower robots to perceive and understand their surroundings with human-like accuracy.
	Furthermore, to foster seamless human-robot collaboration, it is imperative to leverage human-robot interaction technologies.
	This includes harnessing NLP for smooth and intuitive conversations and creating augmented reality (AR)-based collaboration interfaces, thereby ensuring efficient and precise communication and cooperation between humans and robots.
	
	\subsection{Providing Advanced Services via IoRT}
	By interconnecting robots, IoRT facilitates information sharing and collaboration, paving the way for advanced services across various scenarios. For example, in logistics, robots can handle tasks such as locating, packaging, labeling, and transporting items, enabling smart warehousing and delivery.
	In smart manufacturing, seamless connectivity and collaborative work between robots and smart devices can be achieved. 
	In healthcare, IoRT can assist in patient care, facilitating remote medical consultations and surgical support. 
	In smart homes, consumers can use personal robots to perform tasks such as cleaning the house or walking pets, enhancing comfort and convenience. 
	IoRT can also play a critical role in security and surveillance, particularly in crisis management, where robots can offer assistance in dangerous situations, transport goods, and provide aid to victims.

	\section{Conclusions}
	In this paper, an RIS-aided IoRT network was proposed to address the spectrum scarcity, sensing limitations, latency issues, and energy constraints prevalent in multi-robot systems.
	Comprehensive case studies were conducted to demonstrate the benefits of integrating RISs into IoRT systems.
	The simulation results showed that significant performance enhancements could be achieved by optimizing key parameters, including transceiver beamforming, robot trajectories, and RIS coefficients, using advanced techniques such as reinforcement learning and multi-objective optimization. Additionally, the research challenges and promising opportunities for future work in IoRT networks were discussed.
	As the technology continues to mature, it is envisioned that RISs may play a pivotal role in enabling more sophisticated, efficient, and reliable robotic operations, opening a new era of advancements in intelligent and connected systems.


	\bibliographystyle{IEEEtran}
	\bibliography{IEEEabrv,ref}

\begin{thebibliography}{10}
\providecommand{\url}[1]{#1}
\csname url@samestyle\endcsname
\providecommand{\newblock}{\relax}
\providecommand{\bibinfo}[2]{#2}
\providecommand{\BIBentrySTDinterwordspacing}{\spaceskip=0pt\relax}
\providecommand{\BIBentryALTinterwordstretchfactor}{4}
\providecommand{\BIBentryALTinterwordspacing}{\spaceskip=\fontdimen2\font plus
\BIBentryALTinterwordstretchfactor\fontdimen3\font minus
  \fontdimen4\font\relax}
\providecommand{\BIBforeignlanguage}[2]{{%
\expandafter\ifx\csname l@#1\endcsname\relax
\typeout{** WARNING: IEEEtran.bst: No hyphenation pattern has been}%
\typeout{** loaded for the language `#1'. Using the pattern for}%
\typeout{** the default language instead.}%
\else
\language=\csname l@#1\endcsname
\fi
#2}}
\providecommand{\BIBdecl}{\relax}
\BIBdecl
\renewcommand{\BIBentryALTinterwordstretchfactor}{5}

\bibitem{Yara2019Survey}
Y.~Rizk, M.~Awad, and E.~W. Tunstel, ``Cooperative heterogeneous multi-robot
  systems: A survey,'' \emph{{ACM} Comput. Surv.}, vol.~52, no.~2, pp. 1--31,
  2019.

\bibitem{Luo2024TWC}
R.~Luo, H.~Tian, W.~Ni, J.~Cheng, and K.-C. Chen, ``Deep reinforcement learning
  enables joint trajectory and communication in internet of robotic things,''
  \emph{IEEE Trans. Wireless Commun.}, early access, 2024, doi:
  10.1109/TWC.2024.3462450.

\bibitem{Liu2024Reconfigurable}
Z.~Liu, Y.~Liu, and X.~Chu, ``Reconfigurable-intelligent-surface-assisted
  indoor millimeter-wave communications for mobile robots,'' \emph{IEEE
  Internet Things J.}, vol.~11, no.~1, pp. 1548--1557, Jan. 2024.

\bibitem{Shu2024RIS}
J.~Shu, K.~Ota, and M.~Dong, ``{RIS}-enabled integrated sensing, computing, and
  communication for internet of robotic things,'' \emph{IEEE Internet Things
  J.}, vol.~11, no.~20, pp. 32\,503--32\,513, Oct. 2024.

\bibitem{Niotaki2023RF}
K.~Niotaki, N.~B. Carvalho, A.~Georgiadis, X.~Gu, S.~Hemour \emph{et~al.},
  ``{RF} energy harvesting and wireless power transfer for energy autonomous
  wireless devices and {RFIDs},'' \emph{IEEE Journal of Microwaves}, vol.~3,
  no.~2, pp. 763--782, Apr. 2023.

\bibitem{Ren2024Connected}
Y.~Ren, R.~Xie, F.~R. Yu, R.~Zhang, Y.~Wang \emph{et~al.}, ``Connected and
  autonomous vehicles in {Web3}: An intelligence-based reinforcement learning
  approach,'' \emph{IEEE Trans. Intell. Transp. Syst.}, vol.~25, no.~8, pp.
  9863--9877, Aug. 2024.

\bibitem{Luo2022Federated}
R.~Luo, W.~Ni, H.~Tian, and J.~Cheng, ``Federated deep reinforcement learning
  for {RIS}-assisted indoor multi-robot communication systems,'' \emph{IEEE
  Trans. Veh. Technol.}, vol.~71, no.~11, pp. 12\,321--12\,326, Nov. 2022.

\bibitem{Gao2023Intelligent}
X.~Gao, X.~Mu, W.~Yi, and Y.~Liu, ``Intelligent trajectory design for
  {RIS-NOMA} aided multi-robot communications,'' \emph{IEEE Trans. Wireless
  Commun.}, vol.~22, no.~11, pp. 7648--7662, Nov. 2023.

\bibitem{Wang2024Adaptive}
X.~Wang, L.~Nkenyereye, S.~Rani, and J.~Lyu, ``Adaptive sensing for internet of
  robotic things platforms with integrated sensing, computing, and
  communication capabilities,'' \emph{IEEE Internet Things J.}, vol.~11,
  no.~20, pp. 32\,404--32\,415, Oct. 2024.

\bibitem{Ni2022Integrating}
W.~Ni, Y.~Liu, Z.~Yang, H.~Tian, and X.~Shen, ``Integrating over-the-air
  federated learning and non-orthogonal multiple access: What role can {RIS}
  play?'' \emph{IEEE Trans. Wireless Commun.}, vol.~21, no.~12, pp.
  10\,083--10\,099, Dec. 2022.

\bibitem{Luo2023TCOM}
R.~Luo, W.~Ni, H.~Tian, J.~Cheng, and K.-C. Chen, ``Joint trajectory and radio
  resource optimization for autonomous mobile robots exploiting multi-agent
  reinforcement learning,'' \emph{IEEE Trans. Commun.}, vol.~71, no.~9, pp.
  5244--5258, Sept. 2023.

\bibitem{He2023Full}
Z.~He, W.~Xu, H.~Shen, D.~W.~K. Ng, Y.~C. Eldar \emph{et~al.}, ``Full-duplex
  communication for {ISAC}: Joint beamforming and power optimization,''
  \emph{IEEE J. Sel. Areas Commun.}, vol.~41, no.~9, pp. 2920--2936, Sept.
  2023.

\bibitem{Wang2024Multi}
P.~Wang, D.~Han, Y.~Cao, W.~Ni, and D.~Niyato, ``Multi-objective
  optimization-based waveform design for multi-user and multi-target
  {MIMO-ISAC} systems,'' \emph{IEEE Trans. Wireless Commun.}, vol.~23, no.~10,
  pp. 15\,339--15\,352, Oct. 2024.

\bibitem{Zhang2024TGCN}
X.~Zhang, H.~Tian, W.~Ni, Z.~Yang, and M.~Sun, ``Deep reinforcement learning
  for energy efficiency maximization in {SWIPT}-based over-the-air federated
  learning,'' \emph{IEEE Trans. Green Commun. and Netw.}, vol.~8, no.~1, pp.
  525--541, Mar. 2024.

\bibitem{Zhang2024Deep}
X.~Zhang, H.~Tian, W.~Ni, and Z.~Yang, ``Deep reinforcement learning for
  multi-functional {RIS}-aided over-the-air federated learning in internet of
  robotic things,'' in \emph{Proc. IEEE ICC}, Denver, CO, USA, Jun. 2024, pp.
  5461--5466.

\end{thebibliography}

\end{document}